\pdfoutput=1

\documentclass[11pt,a4paper]{article}
\usepackage[hyperref]{acl2023}
\usepackage{times}
\usepackage{inconsolata}
\usepackage{latexsym}
\usepackage{danudefs}
\usepackage{algorithmic}
\usepackage{algorithm}
\usepackage{color}
\usepackage{booktabs}
\usepackage{xspace}
\usepackage{url}
\usepackage{xurl}
\usepackage{braket}

\usepackage{amsmath}
\usepackage[export]{adjustbox}
\usepackage{amssymb}
\usepackage{subfigure}
\usepackage{float}
\usepackage{fdsymbol}
\usepackage{multirow}
\usepackage{setspace}

\usepackage{inconsolata}
\usepackage{microtype}
\usepackage{graphicx}

\usepackage[english]{babel}
\usepackage{hyperref}
\addto\captionsenglish{%
}
\addto\extrasenglish{%
}

\usepackage{acronym}
\acrodef{MLM}[MLM]{Masked Language Model}
\acrodef{SoTA}[SoTA]{state-of-the-art}

\title{Evaluating Short-Term Temporal Fluctuations of Social Biases \\ in Social Media Data and Masked Language Models}

 \author{Yi Zhou$^\spadesuit$ \And Danushka Bollegala$^{\dagger,\clubsuit}$ \\
         Cardiff University$^\spadesuit$, University of Liverpool$^\clubsuit$, Amazon$^\dagger$ \\
         {\tt \{Zhouy131,CamachoColladosJ\}@cardiff.ac.uk} \\
         {\tt danushka@liverpool.ac.uk}
         \And Jose Camacho-Collados$^\spadesuit$   
         }

\date{}

\begin{document}
\maketitle

\begin{spacing}{0.98}

\begin{abstract}
Social biases such as gender or racial biases have been reported in language models (LMs), including Masked Language Models (MLMs).
Given that MLMs are continuously trained with increasing amounts of additional data collected over time, an important yet unanswered question is how the social biases encoded with MLMs vary over time.
In particular, the number of social media users continues to grow at an exponential rate, and it is a valid concern for the MLMs trained specifically on social media data whether their social biases (if any) would also amplify over time.
To empirically analyse this problem, we use a series of MLMs pretrained on chronologically ordered temporal snapshots of corpora.
Our analysis reveals that, although social biases are present in all MLMs, most types of social bias remain relatively stable over time (with a few exceptions). 
To further understand the mechanisms that influence social biases in MLMs, we analyse the temporal corpora used to train the MLMs.
Our findings show that some demographic groups, such as \textit{male}, obtain higher preference over the other, such as \textit{female} on the training corpora constantly.

\end{abstract}

\section{Introduction}
\label{sec:intro}


Despite their usage in numerous NLP applications, MLMs such as BERT~\cite{BERT} and RoBERTa~\cite{liu2019roberta} tend to encode discriminatory social biases expressed in human-written texts in the training corpora~\cite{kurita-etal-2019-measuring,zhou-etal-2022-sense,kaneko-etal-2022-gender}.
For example, if a model is given ``\textit{[MASK] is a nurse.}'' as the input, a gender biased MLM would predict ``\textit{She}'' with a higher likelihood score than for ``\textit{He}'' when filling the [MASK]. Such social biases can result in unfavourable experiences for some demographic groups in certain applications. 
Continuous use of biased models has the potential to amplify biases and unfairly discriminate against users belonging to particular demographic groups.
MLMs are increasingly used in real-world applications such as text generation~\cite{liang-etal-2023-open}, recommendation systems~\cite{malkiel-etal-2020-recobert, kuo2023applying},  search engines~\cite{achsas2022academic, li2023graph} and dialogue systems~\cite{song-etal-2021-bob, park2022bert}.
Therefore, it is crucial to study how MLMs potentially shape social biases. 

On the other hand, social biases may change due to societal changes, cultural shifts and technological advancements. 
MLMs have been trained on ever-increasing massive corpora, often collected from the Web.
In particular, posts on social media, such as but not limited to Reddit and X (former Twitter), have been used to train MLMs.
Social biases contained in the training data are inadvertently learned and perpetuated by MLMs.
At the time of writing, there are 5.07 billion social media users worldwide with 259 million new users joining since this time in 2023.\footnote{\url{https://datareportal.com/social-media-users}}
Given this rapid increase and the significance of social media data as a source for training MLMs, an open question is \textbf{whether LMs trained on social media data continue to demonstrate increasing levels of social biases.}

To answer this question, we investigate multiple MLMs pretrained on snapshots of corpora collected from X at different points in time and evaluate the social biases in those MLMs using multiple benchmark datasets.
We evaluate different types of social biases and observe that the overall bias tends to be stable over time, however, certain types of biases, such as race, skin color, religion, and sexual orientation, exhibit fluctuation over time. 
Based on the experimental results, we note that relying exclusively on the overall bias score can be misleading when evaluating social bias in MLMs, which highlights the importance of evaluating individual bias scores before deploying a model in downstream applications.
Note that we primarily investigate whether language models (LMs) trained on social media data exhibit increasing levels of social biases over time in this paper. 
Our focus is on examining the trends in temporal variations of social biases in both models and datasets. 
Exploring the underlying causes could lead to sociologically oriented experiments and research questions, which are beyond the scope of this NLP-focused study.

\section{Related Work}
\label{sec:related}

\noindent \textbf{Social Biases in NLP.} Social biases in NLP were first drawn to attention by~\citet{bolukbasi2016man}, with the famous analogy “\textit{man is to computer programmer as woman is to homemaker}” provided by static word embeddings.
To evaluate social biases in word embeddings, word Embedding Association Test~\cite[\textbf{WEAT};][]{WEAT} was introduced to measure the bias between two sets of target terms with respect to two sets of attribute terms.
Subsequently, Word Association Test~\cite[\textbf{WAT};][]{du-etal-2019-exploring} was proposed to compute a gender information vector for each word within an association graph~\cite{Deyne2019TheW} through the propagation of information associated with masculine and feminine words.
Follow-up studies investigate social biases in additional models~\cite{liang-etal-2020-towards, liang-etal-2020-monolingual, zhou-etal-2022-sense} and languages~\cite{mccurdy2020grammatical, lauscher-etal-2020-araweat, reusens-etal-2023-investigating, zhou-etal-2023-predictive}.


In contrast, alternative research focuses on social biases in various downstream applications. 
\citet{kiritchenko2018examining} assessed gender and racial biases across 219 automatic sentiment analysis systems, revealing statistically significant biases in several of these systems.
\citet{diaz2018addressing} investigated age-related biases in sentiment classification and found that many sentiment analysis systems, as well as word embeddings, encode significant age bias in their outputs.
\citet{savoldi2021gender} studied gender biases and sentiment biases associated with person name translations in neural machine translation systems. 

Current bias evaluation methods use different approaches, including pseudo-likelihood.~\cite{kaneko2022unmasking}, cosine similarity~\cite{caliskan2017semantics, may-etal-2019-measuring}, inner-product~\cite{ethayarajh2019understanding}, among others.
Independently of any downstream tasks, intrinsic bias evaluation measures~\cite{nangia-etal-2020-crows, nadeem-etal-2021-stereoset, kaneko2022unmasking} assess social biases in MLMs on a standalone basis.
Nevertheless, considering that MLMs serve to represent input texts across various downstream tasks, several prior studies have suggested that the evaluation of social biases should be conducted in relation to those specific tasks~\cite{de2019bias,webster2020measuring}.
\citet{kaneko-bollegala-2021-debiasing} demonstrated that there is only a weak correlation between intrinsic and extrinsic social bias evaluation measures.
In this paper, we use AULA which is an intrinsic measure for evaluating social biases in MLMs.


Various debiasing methods have been proposed to mitigate social biases in MLMs.
\citet{zhao-etal-2019-gender} proposed a debiasing method by swapping the gender of female and male words in the training data.
\citet{webster2020measuring} showed that dropout regularisation can reduce overfitting to gender information, thereby can be used for debiasing pretrained language models.
\citet{kaneko-bollegala-2021-debiasing} proposed a method for debiasing by orthogonalising the vectors representing gender information with the hidden layer of a language model given a sentence containing a stereotypical word.
Our focus in this paper is the evaluation of social biases rather than proposing bias mitigation methods.

\noindent \textbf{Temporal Variations in MLMs.} Diachronic Language Models that capture the meanings of words at a specific timestamp have been trained using historical corpora~\cite{HistBERT, Loureiro:2022}.
\citet{rosin-radinsky-2022-temporal} introduced a temporal attention mechanism by extending the self-attention mechanism in transformers. 
They took into account the time stamps of the documents when calculating the attention scores.
\citet{tang-etal-2023-learning} proposed an unsupervised method to learn dynamic contextualised word embeddings via time-adapting a pretrained MLM using prompts from manual and automatic templates.
\citet{Aida:ACL:2023} proposed a method to predict the semantic change of words by comparing the distributions of contextualised embeddings for the word between two corpora sampled at different timestamps.
\citet{tang2023can} used word sense distributions to predict semantic changes of words in English, German, Swedish and Latin.

On the other hand, \newcite{Zeng2017-rh} learned \emph{socialised} word embeddings by taking into account both the personal characteristics of language used by a social media user and the social relationships of that user.
\newcite{welch-etal-2020-compositional} learned demographic word embeddings, covering attributes such as age, gender, location and religion. 
\newcite{hofmann-etal-2021-dynamic} demonstrated that temporal factors exert a more significant influence than socio-cultural factors in determining the semantic variations of words.
However, to the best of our knowledge, the temporal changes of social biases in MLMs remains understudied, and our focus in this paper is to fill this gap.

\section{Temporal Data and Models}
\label{sec:data_models}

To investigate the temporal variant of social biases appearing in the corpora, we retrieve the posts on X with different timestamps. 
Furthermore, we take into account the MLMs trained on those temporal corpora to study how MLMs potentially shape social biases from these corpora.
In this section, we describe the temporal data and the MLMs that we used in the paper.

\subsection{Temporal Corpora}
\label{sec:corpora}
We use the snapshots of corpora from X across a two-year time span -- from the year 2020 to 2022, collected using Twitter's Academic API.\footnote{Twitter Academic API was interrupted in 2023, and that is the reason why our data collection was interrupted after the end of 2022.}
To obtain a sample that is reflective of the general conversation of people's daily lives on social media, we follow the collection process from \citet{loureiro2022timelms} in order to collect a diverse corpus while avoiding duplicates and spam. 

Specifically, we use the API to retrieve tweets using the most frequently used stopwords,\footnote{We select the top 10 ones from \url{https://raw.githubusercontent.com/first20hours/google-10000-english/master/google-10000-english.txt}} capturing a predetermined number of tweets at intervals of 5 minutes. 
This process is carried out for each hour and every day, spanning a specific quarterly period in the year.
In addition, we leverage specific flags supported by the API to exclusively fetch tweets in English, disregarding retweets, quotes, links, media posts, and advertisements.
Assuming bots are among the most active users, we eliminate tweets from the top $1\%$ of the most frequent posters.

To ensure the dataset remains free of duplicates, we eliminate both exact and near-duplicate tweets.
Specifically, we first convert tweets to lowercase and remove punctuation.
Then we identify near-duplicates by generating hashes using MinHash~\cite{Broder1997} with 16 permutations.
Finally, non-verified user mentions are substituted by a generic placeholder (@user). 
The statistics of temporal corpora can be found in~\autoref{sec:statistics-corpora}.

\subsection{Models trained on Different Timestamps}
\label{sec:models}
To investigate whether social biases in MLMs exhibit temporal variation, we evaluate social biases in MLMs that are trained on corpora sampled from different timestamps. 
Specifically, we select the pre-trained TimeLMs\footnote{\url{https://github.com/cardiffnlp/timelms}}~\cite{loureiro2022timelms}, which are a set of language models trained on diachronic data from X. 
TimeLMs are continuously trained using data collected from X, starting with the initial RoBERTa base model~\cite{liu2019roberta}. 
The base model of TimeLMs is first trained with data until the end of 2019. 
Since then, subsequent models have been routinely trained every three months, building upon the base model.
To ensure the models trained on the corpora sampled with different timestamps are with the same setting (i.e., with incremental updates), we discard the base model trained until 2019 and select the models trained with the temporal corpora described in~\autoref{sec:corpora}. 

To investigate the fluctuations in social biases in MLMs over time, we require a series of pretrained MLMs of the same architecture, trained on corpora sampled at different timestamps. To the best of our knowledge, such MLMs based on architectures other than RoBERTa do not currently exist. Furthermore, training these temporal models from scratch, such as pre-training MLMs with a different architecture, is computationally expensive and time-consuming. For instance, training a RoBERTa base temporal model takes approximately 15 days on 8 NVIDIA V100 GPUs. Given that pretrained temporal MLMs based on models other than RoBERTa are not available, and \citet{zhou-etal-2023-predictive} show that various underlying factors differentially impact social biases in MLMs, our approach focuses on using models that have been continuously trained from an existing RoBERTa base checkpoint. This strategy maintains consistency in model settings, which aids in accurately assessing how MLMs reflect the temporal variations in social biases.

\section{Experimental Setting}
\label{sec:exp}
Our goal in this paper is to study whether MLMs capture temporal changes in social biases, following the same patterns observed in the biases present in training corpora.
For this purpose, we evaluate social biases in MLMs and compare the biases observed in training corpora.

\subsection{Bias Evaluation Metrics}
To investigate the social biases within MLMs, we compute social bias scores of TimeLMs using All Unmasked Likelihood with Attention weights~~\cite[AULA;][]{kaneko2022unmasking}.
This metric evaluates social biases by using MLM attention weights to reflect token significance.
AULA has proven to be more robust against frequency biases in words for evaluating social biases in MLMs and offers more reliable evaluations in comparison to alternative metrics when assessing social biases in MLMs~\cite{Kaneko:2023}.
Further details on the computation of AULA are shown in~\autoref{sec:ap_aula}

\subsection{Benchmarks}
We perform experiments on the two most commonly used benchmark datasets used to evaluate social biases in MLMs.

\noindent \textbf{CrowS-Pairs} \cite{nangia-etal-2020-crows}. proposed Crowdsourced
Stereotype Pairs benchmark (CrowS-Pairs), which is designed to explore stereotypes linked to historically disadvantaged groups.
It is a crowdsourced dataset annotated by workers in the United States and contains nine social bias categories: race, gender, sexual orientation, religion, age, nationality, disability, physical appearance, and socioeconomic status/occupation.
In the CrowS-Pairs dataset, test instances comprise pairs of sentences, where one sentence is stereotypical and the other is anti-stereotypical.
Annotators are instructed to generate examples that indicate stereotypes by contrasting historically disadvantaged groups with advantaged groups.

\noindent \textbf{StereoSet} \cite{nadeem-etal-2021-stereoset}. created StereoSet, which includes associative contexts encompassing four social bias types: race, gender, religion, and profession. 
StereoSet incorporates test instances at both intrasentence and intersentence discourse levels.
They introduced a Context Association Test (CAT) to assess both the language modelling ability and the stereotypical biases of pretrained MLMs.
Specifically, when presented with a context associated with a demographic group (e.g., female) and a bias type (e.g., gender), three distinct labels are provided to instantiate its context, corresponding to a stereotypical, anti-stereotypical, or unrelated association.

We use the social bias evaluation tool released by~\citet{kaneko2022unmasking}\footnote{\url{https://github.com/kanekomasahiro/evaluate_bias_in_mlm}} with its default settings for all evaluations reported in this paper.

\section{Temporal Variation of Social Biases}
In this section, we describe the key findings of our paper, presenting a comprehensive analysis and interpretation of the results.

\subsection{Biases in MLMs}

\begin{figure*}[t!]
    \centering
    \small
    \subfigure[CrowS-Pairs]{
       
        \begin{minipage}[b]{0.5\linewidth}
        \includegraphics[width=0.9\linewidth]{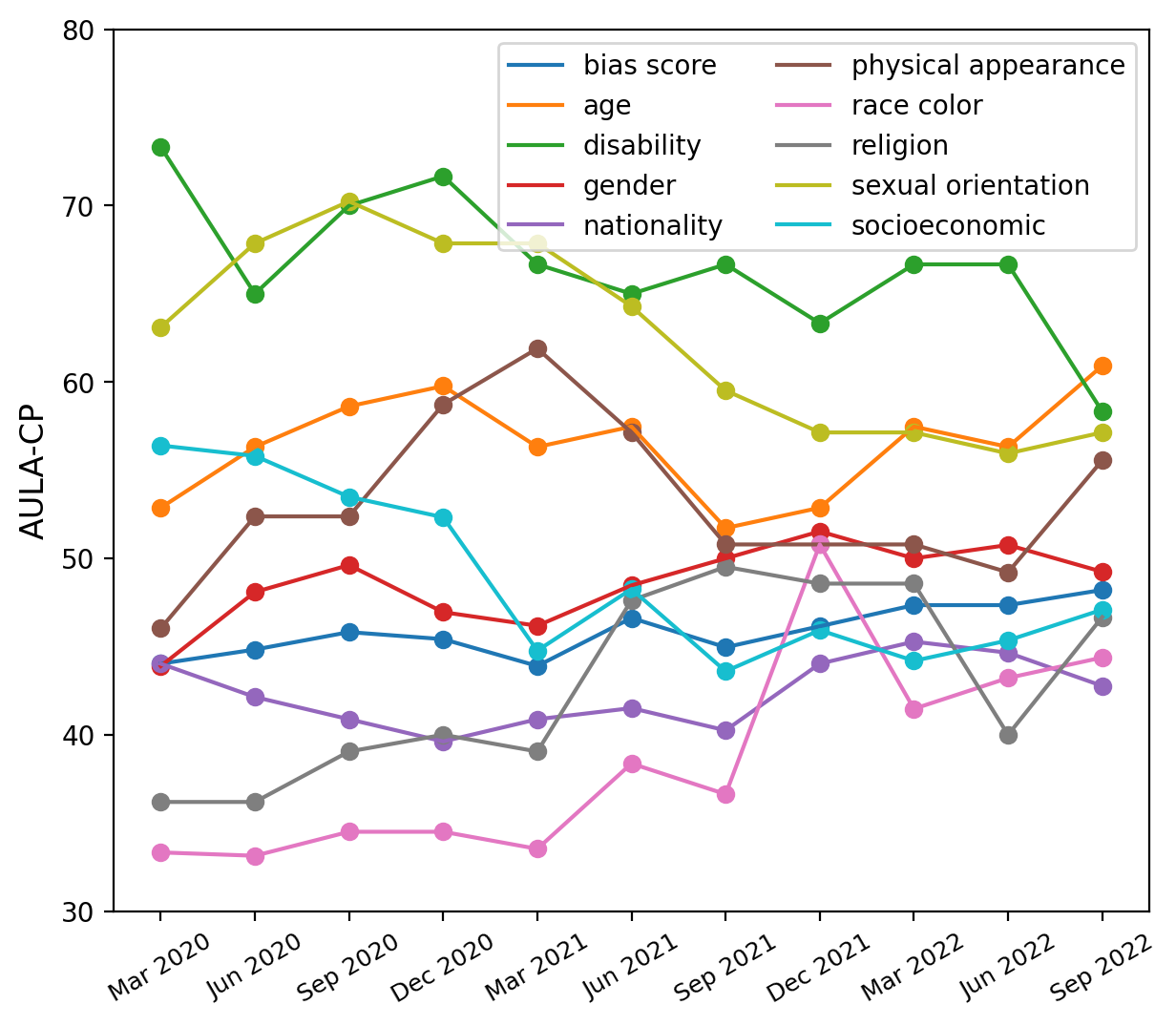}

        \end{minipage}
    }
    \subfigure[StereoSet]{
        \begin{minipage}[b]{0.45\linewidth} 
        \includegraphics[width=1.0\linewidth]{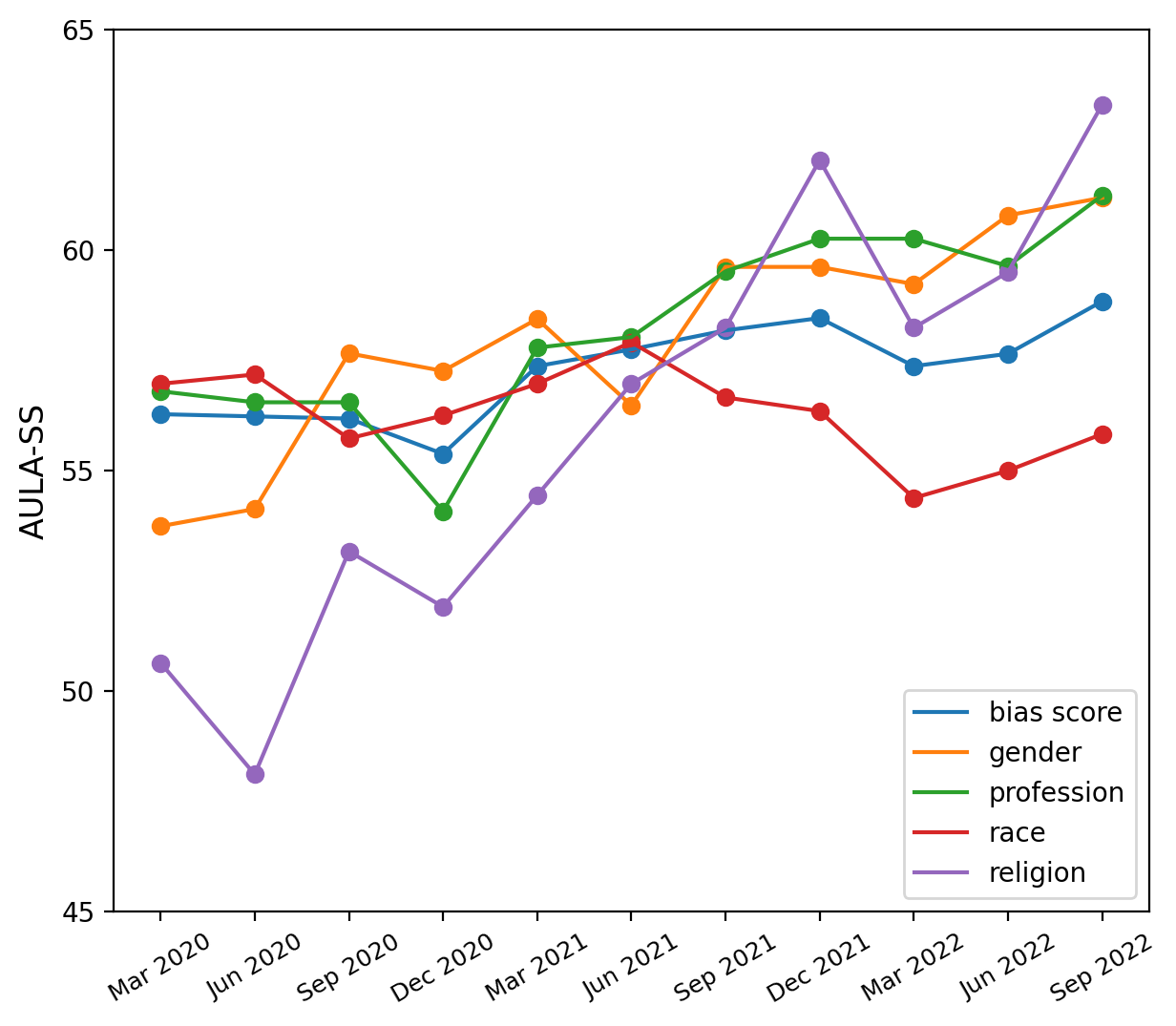}
       \label{fig:aula-scores-ss}
       \vspace{-0.5cm}
        \end{minipage}
    }
    \caption{Social bias scores across time for different types of biases computed using the AULA metric. Results evaluated on the CrowS-Pairs and StereoSet datasets are shown respectively on the left and right. The `bias score' (in dark blue) indicates the overall bias score.}
    \label{fig:aula-scores}
    \vspace{-1.5mm}
\end{figure*}

\autoref{fig:aula-scores} shows the changes of bias scores for different bias types in TimeLMs over the period from March 2020 to September 2022 computed by AULA on both CrowS-Pairs and StereoSet datasets.
It is noticeable that different types of biases within TimeLMs change over time.
The overall bias scores exhibit minimal changes over time compared to other types of biases in both datasets. 
This result suggests that even when there is no overall social bias reported by a metric, an MLM can still be biased with respect to a subset of the bias types. Therefore, it is important to carefully evaluate bias scores per each bias type before an MLM is deployed in downstream applications. 


\begin{table}[hbt!]
    \centering
    \resizebox{0.48\textwidth}{!}{
    \begin{tabular}{lcccc}
    \toprule
       &mean & lower/upper & SE &SD  \\
        \midrule
        \textbf{CrowS-Pairs} & & & & \\
       OVERALL BIAS &45.88 & 45.21/46.55 & 0.41 &1.41 \\
       race-color &38.53 & 36.19/41.88 & 1.68 &5.77 \\
       sexual-orientation &62.55 & 60.06/65.15 & 1.54 &5.36 \\
       religion &42.86 & 40.35/45.45 & 1.52 &5.30 \\
       socioeconomic &48.84 & 46.78/51.32 & 1.37 & 4.79 \\
       appearance &53.25 & 51.23/55.70 & 1.33 &4.62 \\
       disability &66.67 & 64.70/68.49 & 1.17 &4.08 \\
       age &56.42 & 54.86/57.68 & 0.85 &2.93 \\
       gender &48.61 & 47.40/49.55 & 0.64 &2.23 \\
       nationality &42.37 & 41.51/43.28 & 0.55 &1.91 \\
       \midrule
        \textbf{StereoSet} & & & & \\
        OVERALL BIAS &57.23 & 56.70/57.74 & 0.31 &1.09 \\
       religion &56.04 & 53.62/58.34 & 1.39 &4.81 \\
       gender &58.00 & 56.72/59.07 & 0.71 &2.47 \\
       profession &58.24 & 57.15/59.22 & 0.62 &2.15 \\
       race &56.28 & 55.77/56.73 & 0.29 &1.02 \\
        \midrule
    \end{tabular}
    }
    \caption{Confidence intervals and standard errors are computed using bootstrapping test for each bias type on the CrowS-Pairs and StereoSet benchmarks. SE and SD represent standard error and standard deviation, respectively. Lower/upper indicates the lower/upper bound of the confidence intervals. In each dataset, different bias types are sorted in the descending order of their SD. }
    \label{tab:bootstrap-cp}
\end{table}

When evaluating on CrowS-Pairs, we observe that both disability and sexual orientation biases consistently receive bias scores above 50. 
This indicates a consistent inclination of these two biases toward stereotypical examples over a span of two years.
Conversely, religion and nationality exhibit a consistent inclination toward anti-stereotypical examples over time.
In terms of the evaluation on StereoSet, most types of biases exhibit stereotypical tendencies, except the religious bias in June 2020, which leaned toward anti-stereotypical examples.
In particular, the religious biases have increased from 51 to 63 over the two year period from 2020 to 2022.
This finding highlights the nuanced nature of different types of biases and their variations across different contexts, encouraging future research aimed at establishing a benchmark that equally considers different types of biases~\cite{blodgett-etal-2021-stereotyping}.
However, our primary focus is on investigating the temporal fluctuations of social biases in MLMs, and as such, the specific direction of different biases presenting differently on the evaluation datasets is out of the scope of this paper.

       

\begin{figure*}[hbt!]
    \centering
    \subfigure[CrowS-Pairs]{
        \begin{minipage}[b]{0.5\linewidth}
        \includegraphics[width=0.9\linewidth]{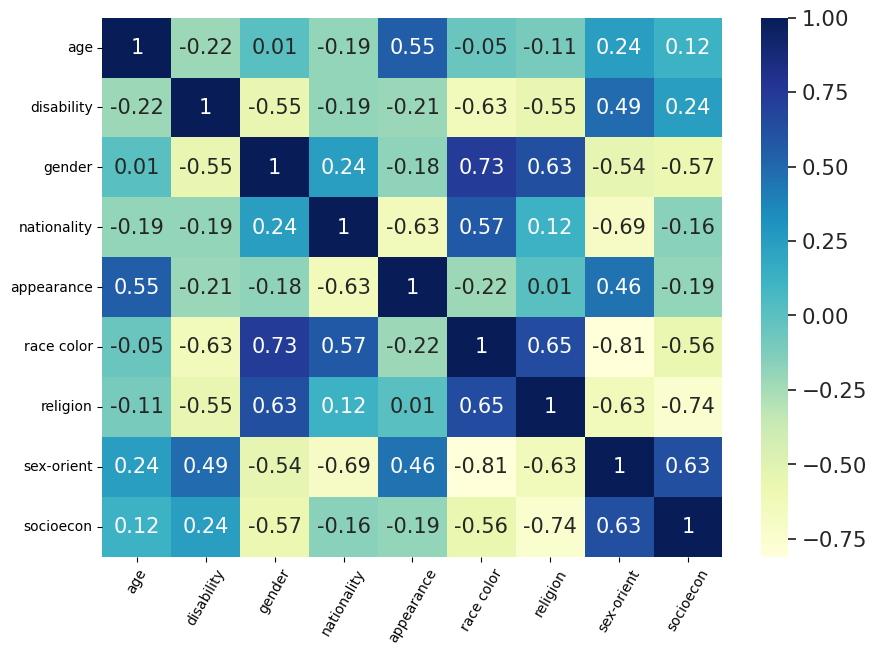}
        \end{minipage}
    }
    \subfigure[StereoSet]{
        \begin{minipage}[b]{0.45\linewidth} 
        \includegraphics[width=1.0\linewidth]{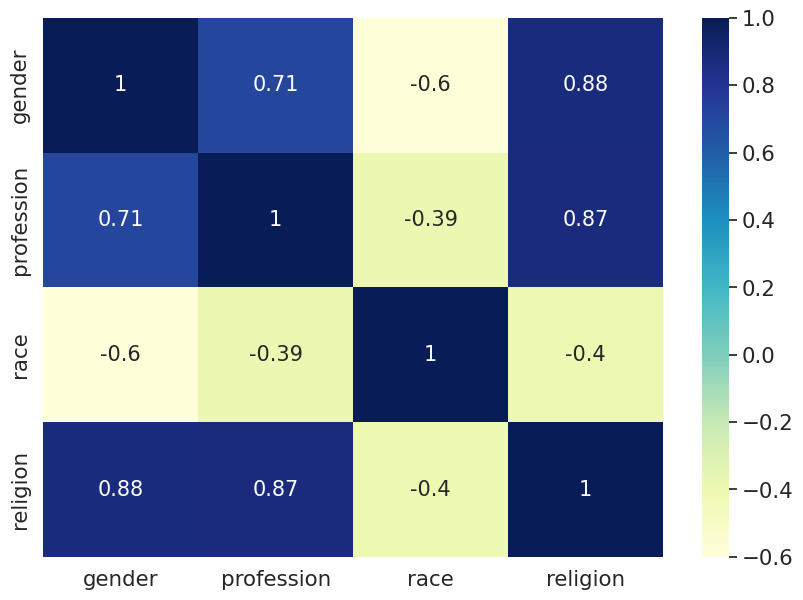}
        \end{minipage}
    }
    \caption{Pearson correlation coefficient of each pair of bias types. Results on the CrowS-Pairs and StereoSet datasets are shown respectively on the left and right.}
    \label{fig:correlation}
    \vspace{-1.5mm}
\end{figure*}

\noindent \textbf{Statistical indicators of bias fluctuation changes.}
To further validate the consistency of the aforementioned observations, we use the bootstrapping significance test~\cite{tibshirani1993introduction} to the temporal variation of different social bias types.
Specifically, given a bias type, we first compute the AULA score over the entire dataset at a particular time point, resulting in a series of data points, each one corresponding to a particular time point, and we report the average and standard deviation of that score along with its confidence interval and standard error computed using bootstrapping.
Bootstrapping is a statistical technique which uses random sampling with replacement. 
By measuring the properties when sampling from an approximating distribution, bootstrapping estimates the properties of an estimand (e.g., variance).  
We implement bootstrapping using the SciPy\footnote{\url{https://docs.scipy.org/doc/scipy/reference/generated/scipy.stats.bootstrap.html}} at 0.9 confidence level to compute the confidence intervals, while setting other parameters to their defaults. 

\autoref{tab:bootstrap-cp} shows the result. 
In CrowS-Pairs, the bias types such as sexual orientation, physical appearance, disability, and age manifest biases mostly toward stereotypical examples (i.e., the mean of their bias scores are above 50), while biases associated with race colour, religion, socioeconomic, gender and nationality tend to have biases toward anti-stereotypical examples (i.e., the mean of their bias scores are below 50).
On the other hand, race colour reports the highest standard error, indicating that it is the most fluctuating bias type over time. 

In StereoSet, we observe all the types of biases exhibit biases toward stereotypical examples. 
Moreover, religion is the most fluctuating bias over time compared to other types of biases, while racial bias does not change much over time. 
Note that the CrowS-Pairs dataset assesses race colour bias, specifically concentrating on the skin colour associated with race, which is different from the race bias considered in StereoSet.

\begin{figure*}[hbt!]
    \centering
    \subfigure[Gender]{
    \label{fig:gender}
        \begin{minipage}[]{0.45\linewidth}
        \centerline{\includegraphics[scale=0.4]{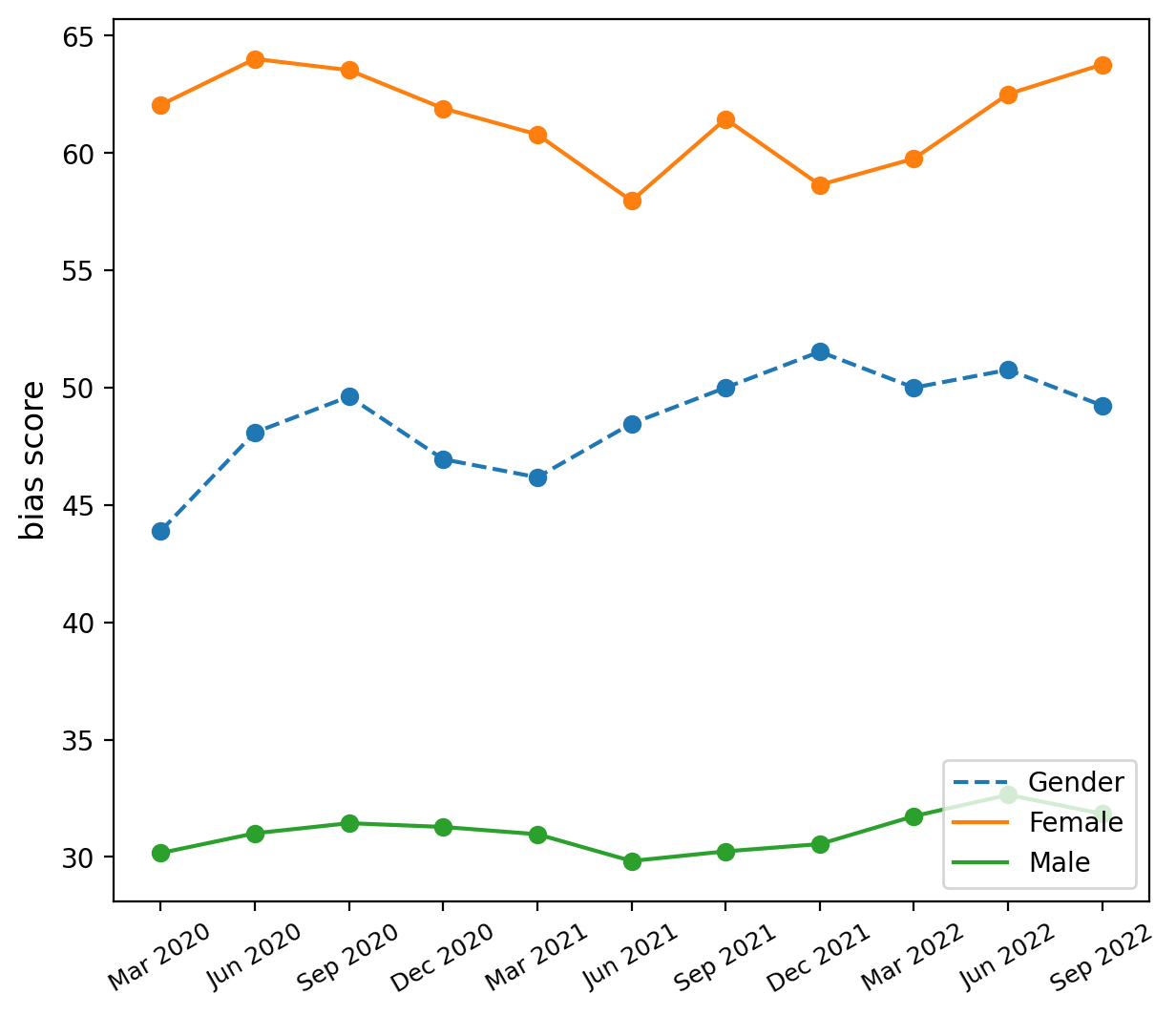}}
        \end{minipage}
    }
    \subfigure[Race]{
    \label{fig:race}
        \begin{minipage}[]{0.45\linewidth} 
        \centerline{\includegraphics[scale=0.4]{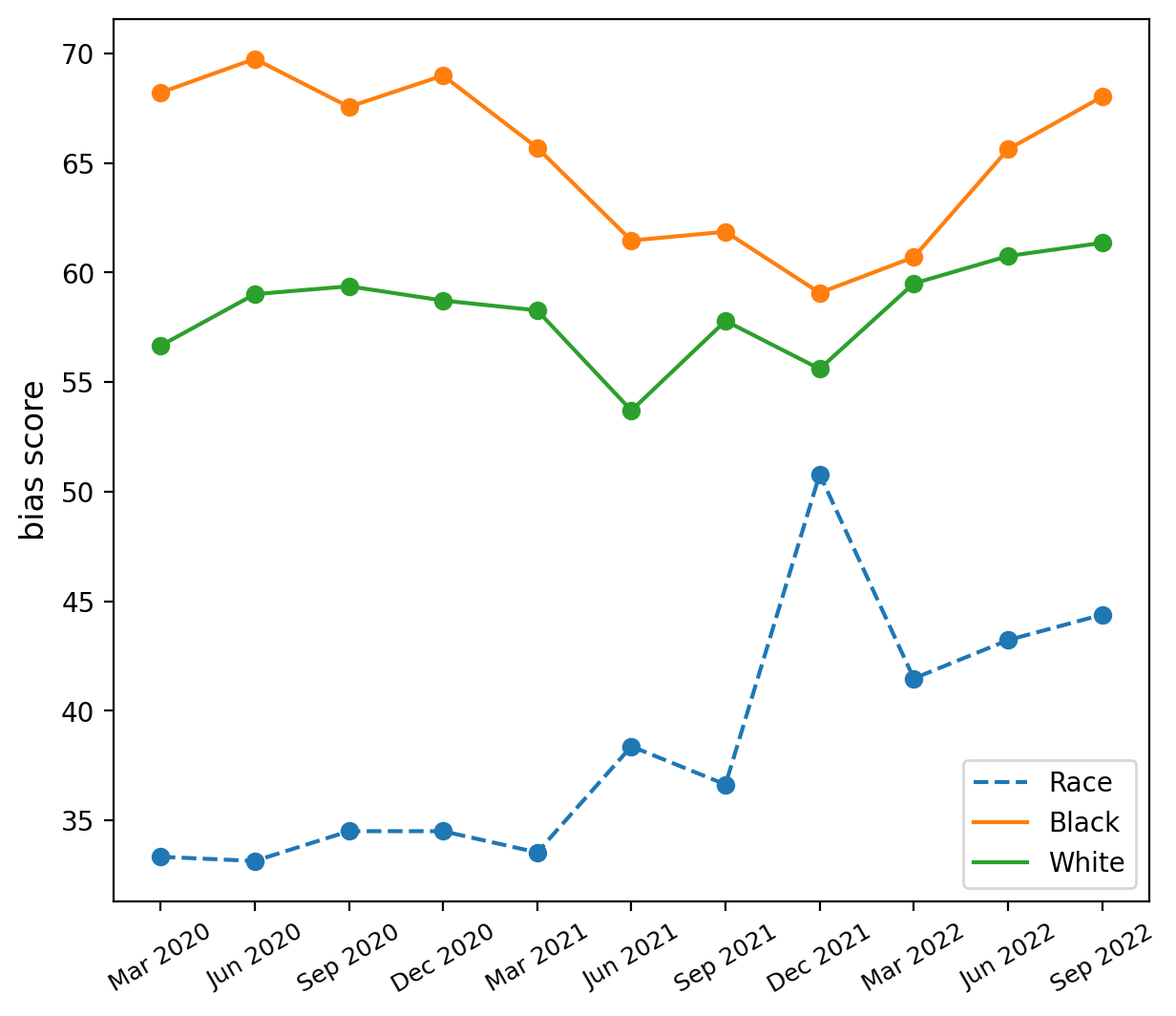}}
        \end{minipage} 
    }
    \\
    \subfigure[Religion]{
    \label{fig:religion}
        \begin{minipage}[]{0.45\linewidth} 
        \centerline{\includegraphics[scale=0.4]{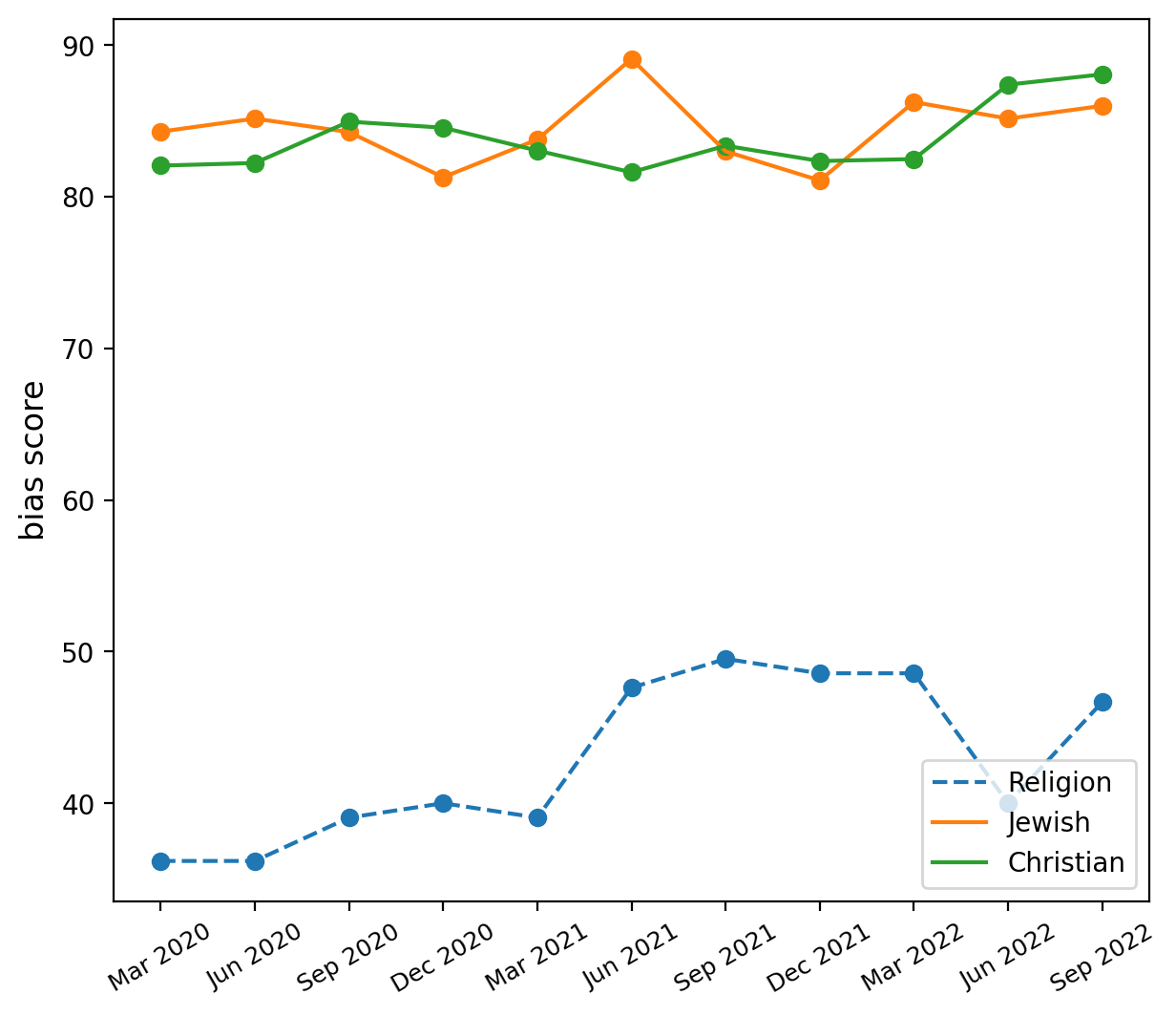}}
        \end{minipage} 
    }
    \subfigure[Age]{
    \label{fig:age}
        \begin{minipage}[]{0.45\linewidth} 
        \centerline{\includegraphics[scale=0.4]{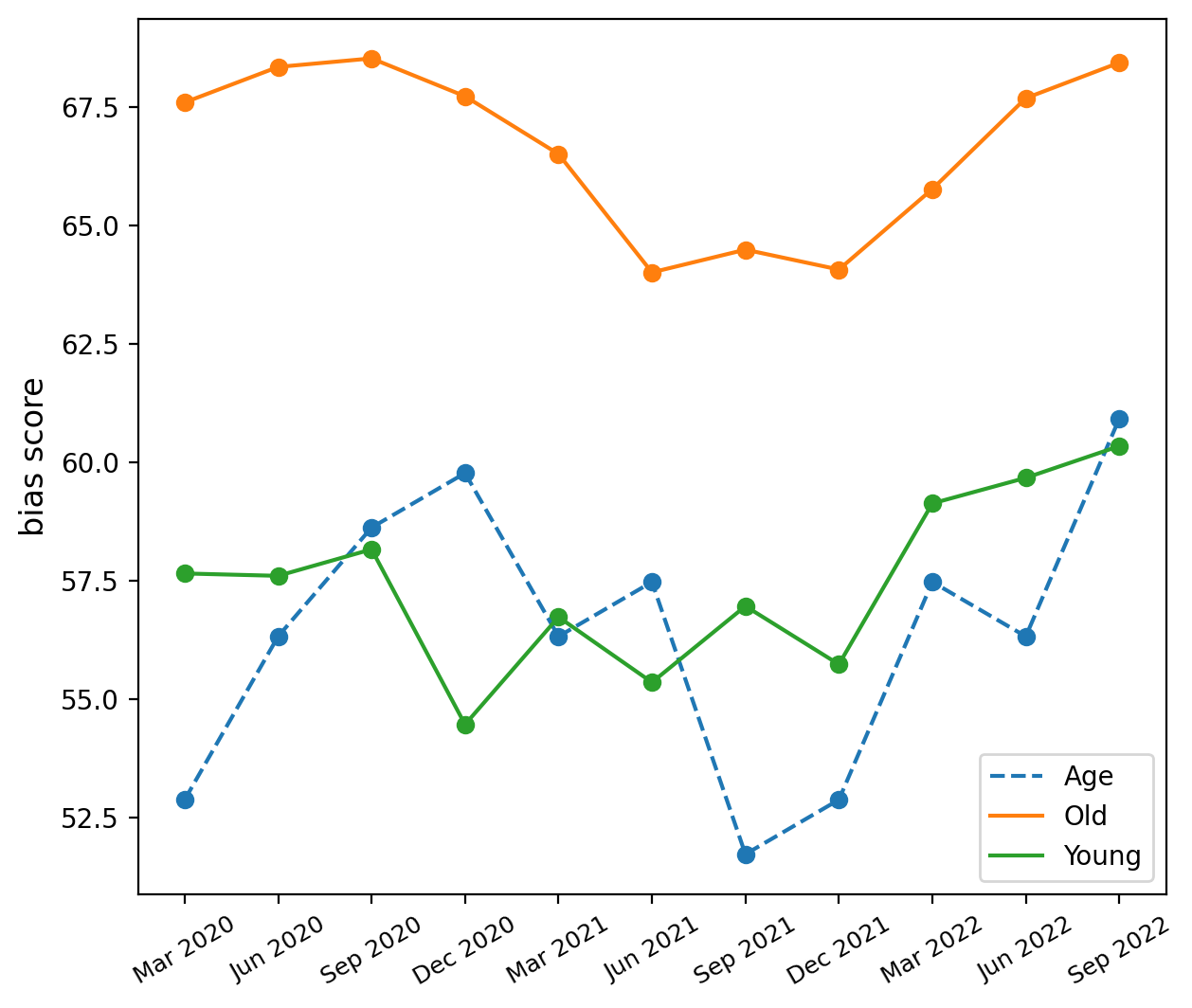}}
        \end{minipage} 
    }
    \caption{Social biases in data associated with different demographic groups. A sentiment classifier is used to determine whether a tweet associated with a particular demographic group conveys positive or negative sentiment. Dash line represents the bias scores computed using \eqref{eq:AULA} on CrowS-Pairs, while solid lines show bias scores computed using \eqref{eq:bias-data}, respectively.}
    \label{fig:bias-data}
    \vspace{-1.5mm}
\end{figure*}



\subsection{Correlations between Bias Types}
\label{sec: correlation}
To investigate whether the change in one type of bias influences other types, we compute the Pearson correlation coefficient ($r$) for each pair of bias types.
We use the SciPy library\footnote{\url{https://docs.scipy.org/doc/scipy/reference/generated/scipy.stats.pearsonr.html}} with the default setting for doing so and show the results in~\autoref{fig:correlation}.
When evaluating on CrowS-Pairs, race color and gender biases have the highest correlation (i.e., 0.73) compared to other bias pairs, whereas race color obtains the lowest correlation (i.e., -0.81) with sexual orientation. 
Moreover, strong positive correlations (i.e., $r>0.65$) exist among pairs such as race colour vs. gender and race colour vs, religion, while sexual orientation vs. race colour, sexual orientation vs. nationality and socioeconomic vs. religion obtain strong negative correlation (i.e., $r<-0.65$).

As far as StereoSet is concerned, we observe that the pairs such as profession vs. gender, religion vs, gender, and religion vs, profession exhibit strong positive correlations (i.e., $r>0.65$), while race vs. gender, race vs. profession, as well as religion vs. race, manifest negative correlations. 

\subsection{Biases in Data}
\label{sec:bias-in-data}

To study the presence of biases related to a certain demographic group in the training corpus and the extent to which an MLM learns these biases during pre-training, we measure different types of social biases appearing in the corpus.
Following prior work that evaluates bias in words using their association to pleasant vs. unpleasant words~\cite{WEAT,du-etal-2019-exploring}, we evaluate the bias score of a demographic group $\cD$ by considering its members $x \in \cD$, and their association with positive and negative contexts. 

However, instead of relying on a fixed set of pleasant/unpleasant words, which is both limited and the occurrence of a single word could be ambiguous, we use sentiment classification as a proxy for eliciting such pleasant (expressed by a positive sentiment) and unpleasant (expressed by a negative sentiment) judgements. 
For this purpose we use the sentiment classification model fine-tuned on TweetEval \cite{barbieri-etal-2020-tweeteval},\footnote{\url{https://huggingface.co/cardiffnlp/twitter-roberta-base-sentiment-latest}} which associates each tweet with a positive, negative or neutral sentiment.
According to \citet{kiritchenko2018examining}, some sentiment analysis models show biases, particularly related to race more than gender. In this paper, we specifically focus on evaluating biases using a state-of-the-art sentiment analysis model, according to the TweetEval benchmark, that has been fine-tuned on tweets to minimise biases that could arise from varied datasets. It is important to note that our analysis does not extend to comparing biases across different sentiment analysis models, which is beyond the scope of this paper.

Given a word $x \in \cD$ that occurs in a sentence $S$, we use the \emph{negativity score} to measure the social biases in the training data. 
The negativity score of the group $\cD$ is defined by \eqref{eq:bias-data}.
\begin{equation}
\small
\label{eq:bias-data}
    {\rm Score} = 100\times\frac{\sum_{x\in \cD}S_{\rm n}(x)}{\sum_{x \in \cD}S_{\rm p}(x)+S_{\rm n}(x)}
\end{equation}
Here, $S_{\rm p}(x)$ and $S_{\rm n}(x)$ represent the number of times that $S$ is classified as respectively \textit{positive} or \textit{negative} by a sentiment classifier given the word $x$ appear in the sentence $S$. 
Similar to the bias score computed using AULA, an unbiased dataset will return a bias score of 50, while greater and lower than 50 indicates the bias toward stereotypical and anti-stereotypical examples, respectively.

We select four types of biases and categorise them according to the magnitude of changes over time. 
Based on the results shown in~\autoref{tab:bootstrap-cp}, we focus on those with minimal changes (i.e., standard error less than 1.00), which are age and gender biases, and those with more pronounced changes (i.e., standard error greater than 1.00), which are race colour and religion for evaluation.
Note that the racial and religious biases in CrowS-Pairs and StereoSet are sub-categorised and cover more than two demographic groups. 
However, in the following evaluation, we take into account two demographic groups for each of the bias types. 

\noindent \textbf{Gender Bias.} We retrieve the top-50 \textit{male} and \textit{female} names respectively from Name Census: United States Demographic,\footnote{\url{https://namecensus.com/baby-names/}} which contains the most popular baby names from 1880 to the latest available data in 2022. 
These names are directly sourced from Social Security card applications submitted for the births in the United States.
The detailed list of the words we used for the demographic descriptor words for gender bias can be found in~\autoref{sec:gender_word_list}.

Figure \autoref{fig:gender} shows the results. 
The \textit{male} category consistently obtains a low negativity score (i.e., $<35$), while \textit{female} returns high negativity scores (i.e., $>55$) across time.
This indicates that the words in the \textit{male} group constantly exhibit a strong association with positive tweets compared to the \textit{female} group. 
Moreover, the \textit{male} bias exhibits stability over time, whereas \textit{female} bias shows more fluctuations. 

\noindent \textbf{Racial Bias.}
To evaluate racial bias occurring in training corpora, we select the names that are associated with being African American and European American from the work by~\citet{kiritchenko2018examining}, consisting of 20 names in each of the demographic groups. 
The lists of words representing \textit{White} and \textit{Black} races used in our paper are shown in~\autoref{sec:race_words}.

From Figure~\autoref{fig:race} we observe that both \textit{Black} and \textit{White} biases reduce from June 2020 to June 2021, while both increase from December 2021 to September 2022.
Conversely, the overall racial bias contains a different trend. 
The overall racial bias remains stable until March 2021. 
In addition, both \textit{Black} and \textit{White} biases have higher levels of social biases toward stereotypical examples, while the overall racial bias tends to be anti-stereotypical, except in December 2021, when it reaches its peak. 

\noindent \textbf{Religious Bias.}
In terms of religious bias, we consider the terms associated with \textit{Jewish} and \textit{Christian} identities and choose terms listed as the demographic identity labels from AdvPromptSet~\cite{esiobu2023robbie}, and the phrases related to demographic groups are listed in~\autoref{sec:religion_words}.

The result of the religious bias scores as well as the negativity scores associated with \textit{Christian} and \textit{Jewish} identities are shown in Figure~\autoref{fig:religion}.
Regarding biases associated with \textit{Jewish} and \textit{Christian} in the data, we observe that both biases obtain high levels of social bias toward stereotypes.
However, the general religious bias in MLMs demonstrates a lower degree of social biases, primarily towards anti-stereotypes over time.
On the other hand, the \textit{Christian} bias is more stable compared to \textit{Jewish} and overall religious biases. 

\noindent \textbf{Age Bias.} 
For the age bias, we consider the demographic categories of \textit{young} and \textit{old}. 
Therefore, we use the descriptor terms in HOLISTICBIAS~\citet{smith-etal-2022-im}, and the list of the terms associated with young and old can be found in~\autoref{sec:age_words}.

Figure~\autoref{fig:age} shows the bias associated with \textit{young} and \textit{old} demographic groups along with the overall age bias over time. 
We observe that from December 2021 to March 2022, the negativity score associated with the \textit{old} group increases along with the overall age bias. However, we can observed a marked difference in terms of absolute values, with the negativity score for the \textit{old} group being generally much larger.

\noindent \textbf{Control Analysis.} 
To further verify whether social biases also vary independently of time, we conduct a control analysis by randomly sampling a subset of a corpus within the same time period.
Specifically, we consider social biases associated with \textit{female} and \textit{male} and randomly sample 1/5 of the tweets from January to March 2020 for 5 times and compute the standard deviation of \textit{female} and \textit{male} bias scores over these samples.

The standard deviations of both \textit{female} and \textit{male} biases in a corpus sampled with the same timestamp are 0.16 and 0.19, respectively, which are much lower than the standard deviations of \textit{female} (i.e., 2.03) and \textit{male} biases (i.e., 0.84) across time. 
This indicates that the temporal aspect has a more pronounced effect on social biases, showing that social biases do not vary independently of time. 
The details of the results for social biases in random sample subsets and in the temporal corpora are shown in~\autoref{sec:ap-control}.



\subsection{Comparison with temporal bias fluctuations in historical data}
To further investigate the fluctuations of social biases present in corpora with a longer time span, we apply the same experimental setting as in~\autoref{sec:exp} on COHABERT,\footnote{\url{https://github.com/seongmin-mun/COHABERT}} which is a series of RoBERTa base models that are continuously trained on COHA~\cite{DVN_2015}.
COHA is the largest structured corpus of historical English. 
The COHABERT models have been trained over a long period, spanning from the year 1810 to 2000.

Due to space limitations, the results for different bias types and their historical fluctuations are shown in the appendix (\autoref{sec:cohabert-bias} and \autoref{sec:cohabert-statistics-bias}, respectively).
Overall, biases show more fluctuations over a longer time span (i.e., exhibiting higher standard deviations over time) than over a shorter one.
Comparing the different bias types within COHABERT models, we observe a similar trend over time, demonstrating that overall bias scores remain relatively stable compared to specific bias types across both CrowS-Pairs and StereoSet. 
Specifically, the overall bias in COHA produced standard deviations of 1.11 in StereoSet and 3.59 in CrowS-Pairs when measured in 10-year span periods.
Sexual orientation is the most fluctuating bias type in CrowS-Pairs, whereas religion shows the most variability over time in StereoSet.

\section{Conclusion}
We studied the temporal variation of social biases appearing in the data as well as in MLMs. 
We conducted a comprehensive study using various pretrained MLMs trained on different snapshots of datasets collected at different points in time.
While social biases associated with some demographic groups undergo changes over time, the results show that the overall social biases, as captured by language models and as analysed on the underlying corpora, remain relatively stable. 
Therefore, using the overall bias score without considering different bias types to indicate social biases present in MLMs can be misleading. 
We encourage future research to consider different types of biases for study, where these biases can be more pronounced.


\section{Limitations}
This paper studies the temporal variation of social biases in datasets as well as in MLMs.
In this section, we highlight some of
the important limitations of this work. 
We hope this will be useful when extending our work in the future by addressing these limitations.

As described in~\autoref{sec:models}, our main results are based on the RoBERTa base models trained with temporal corpora. This is limited by the availability of language models trained on different time periods.
Related to this, the evaluation in this paper is limited to the English language and we only collect temporal corpora on X. 
Extending the work to take into account models with different architectures for comparison and the study to include multiple languages as well as collecting data from different social media platforms will be a natural line of future work. 

As mentioned in \autoref{sec:bias-in-data}, certain sentiment analysis models exhibit biases.
These biases in such models are more commonly found in relation to race compared to gender. 
In this paper, we measure biases in data by only taking into account one RoBERTa based sentiment analysis model trained on tweets. 
However, comparing biases in different sentiment analysis models is out of the scope of this paper.

In this paper, we narrow down our focus to evaluate the intrinsic social biases captured by MLMs.
However, there are various extrinsic bias evaluation datasets existing such as BiasBios~\cite{de2019bias}, STS-bias~\cite{webster2020measuring}, NLI-bias~\cite{dev2020measuring}. 
A logical next step for our research would be to extend our work and assess the extrinsic biases in MLMs.

Due to the computational costs involved when training MLMs, we conduct a control experiment to investigate whether social biases vary independently of time with the focus on biases in data.
However, it remains to be evaluated whether the similar trend can be observed for the biases in MLMs.

\section{Ethical Considerations}
In this paper, we aim to investigate whether social biases in datasets and MLMs exhibit temporal variation.
Although we used datasets collected from X, we did not annotate nor release new datasets as part of this research.
Specifically, we refrained from annotating any datasets ourselves in this study. Instead, we utilised corpora and benchmark datasets that were previously collected, annotated, and consistently employed for evaluations in prior research.
To the best of our knowledge, no ethical issues have been reported concerning these datasets. All the data utilised from X has been anonimized, excluding all personal information and only retaining the text in the post, where user mentions were also removed.

The gender biases considered in the bias evaluation datasets in this paper only consider binary gender.
However, non-binary genders are severely lacking representation in the textual data used for training MLMs~\cite{dev2021harms}.
Moreover, non-binary genders are frequently associated
with derogatory adjectives. 
It is crucial to evaluate social bias by considering non-binary gender.

\end{spacing}

\bibliography{myrefs.bib}
\bibliographystyle{acl_natbib}

\appendix

\section{Statistics of Temporal Corpora Collected from X}
\label{sec:statistics-corpora}
The statistics of temporal corpora collected from X using Twitter's Academic API across a two-year time span (i.e., from the year 2020 to 2022) can be found in \autoref{tab:data_statistics}

\begin{table}[hbt!]
    \centering
    \resizebox{0.48\textwidth}{!}{
    \begin{tabular}{cccc}
        \toprule
        Quarter  & 2020 & 2021 & 2022 \\
        \midrule
        Q1 & 7,917,521 & 9,346,385 & 18,708,819 \\
        Q2 & 7,922,090 & 9,074,847 & 18,536,812 \\
        Q3 & 7,839,401 & 9,388,844 & 18,347,979 \\
        Q4 & 7,769,658 & 9,471,075 & 18,427,616 \\
         Total  & 31,448,670 & 37,281,151 & 74,021,226 \\
         \midrule
    \end{tabular}
    }
    \caption{The statistics of temporal corpora collected from X. Each quarter corresponds to three months. Q1: January-March, Q2: April-June, Q3: July-September, Q4: October-December.}
    \label{tab:data_statistics}
     \vspace{-1.5mm}
\end{table}


\section{All Unmasked Likelihood with Attention (AULA)}
\label{sec:ap_aula}
We compare the pseudo-likelihood scores returned by an MLM for stereotypical and anti-stereotypical sentences using AULA.
This metric evaluates social biases by using MLM attention weights to reflect token significance.

Given a sentence $S = s_1, \dots, s_n$ encompassing a sequence of tokens $s_i$ with a length of $|N|$,
we calculate the Pseudo Log-Likelihood, denoted as $\mathrm{PLL}(S)$, to predict all tokens within sentence $S$, excluding the start and end tokens of the sentence.
The score $\mathrm{PLL}(S)$ for sentence $S$ given by~\eqref{eq:AULA} can be used to assess the preference expressed by an MLM for the given sentence $S$.
\begin{align}
\small
    \label{eq:AULA}
    \mathrm{PLL}(S) \coloneqq \frac{1}{|N|} \sum_{i=1}^{|N|} \alpha_i \log P(s_i | S; \theta)
\end{align}
where $\alpha_i$ is the average of multi-head attention weights associated with each token $s_i$.
$P(s_i | S; \theta)$ indicates the probability of the MLM assigning token $s_i$ given the context of sentence $S$.
The fraction of sentence pairs where the MLM's preference for stereotypical ($S^{st}$) sentences over anti-stereotypical ($S^{at}$) ones is computed as the AULA bias score of the MLM as in~\eqref{eq:score}.
\par\nobreak
{\small
\begin{align}
    \label{eq:score}
  \mathrm{AULA} =  \frac{100}{M}\sum_{(S^{\rm st}, S^{\rm at})} \mathbb{I}(\mathrm{PLL}(S^{\rm st}) > \mathrm{PLL}(S^{\rm at}))
\end{align}
}
Here $M$ denotes the overall count of sentence pairs in the dataset and $\mathbb{I}$ represents the indicator function that yields 1 when its condition is true and 0 otherwise.
The AULA score calculated by~\eqref{eq:score} lies in the interval [0, 100].
An unbiased model would yield bias scores close to 50, while bias scores lower or higher than 50 indicate a bias towards the anti-stereotypical or stereotypical group, respectively.

\section{Demographic Descriptor Words for Biases}

\subsection{Gender Bias}
\label{sec:gender_word_list}
The names associated with female and male for gender biases are listed in \autoref{tab:female-words}.

\begin{table*}
    \centering
    \resizebox{1.0\textwidth}{!}{
    \begin{tabular}{l p{11.5cm}}
    \toprule
        Demographic Group & Terms \\
        \midrule
        Female & Olivia, Emma, Charlotte, Amelia, Sophia, Isabella, Ava, Mia, Evelyn, Luna, Harper, Camila, Sofia, Scarlett, Elizabeth, Eleanor, Emily, Chloe, Mila, Violet, Penelope, Gianna, Aria, Abigail, Ella, Avery, Hazel, Nora, Layla, Lily, Aurora, Nova, Ellie, Madison, Grace, Isla, Willow, Zoe, Riley, Stella, Eliana, Ivy, Victoria, Emilia, Zoey, Naomi, Hannah, Lucy, Elena, Lillian\\
         \midrule
        Male & Liam, Noah, Oliver, James, Elijah, William, Henry, Lucas, Benjamin, Theodore, Mateo, Levi, Sebastian, Daniel, Jack, Michael, Alexander, Owen, Asher, Samuel, Ethan, Leo, Jackson, Mason, Ezra, John, Hudson, Luca, Aiden, Joseph, David, Jacob, Logan, Luke, Julian, Gabriel, Grayson, Wyatt, Matthew, Maverick, Dylan, Isaac, Elias, Anthony, Thomas, Jayden, Carter, Santiago, Ezekiel, Charles\\
        \midrule
    \end{tabular}
    }
    \caption{The words that we used that are associated with female for evaluating gender bias in the corpus.}
    \label{tab:female-words}
\end{table*}

\subsection{Race Bias}
\label{sec:race_words}
The names associated with two different demographic groups for race bias are listed in~\autoref{tab:race_words_list}.

\begin{table*}
    \centering
    \resizebox{1.0\textwidth}{!}{
    \begin{tabular}{l p{11.5cm}}
    \toprule
      Demographic Group & Terms \\
    \midrule
    African American & Ebony, Jasmine, Lakisha, Latisha, Latoya, Nichelle, Shaniqua, Shereen, Tanisha, Tia, Alonzo, Alphonse, Darnell, Jamel, Jerome, Lamar, Leroy, Malik, Terrence, Torrance \\
    
     \midrule
    European American & Amanda, Betsy, Courtney, Ellen, Heather, Katie, Kristin, Melanie, Nancy, Stephanie, Adam, Alan, Andrew, Frank, Harry, Jack, Josh, Justin, Roger, Ryan \\
        \midrule
    \end{tabular}
    }
    \caption{The lists of words representing different demographic groups related to race bias.}
    \label{tab:race_words_list}
\end{table*}

\subsection{Religion Bias}
\label{sec:religion_words}

The terms associated with two different demographic groups for religion bias are listed in~\autoref{tab:religion_word_list}.

\begin{table*}
    \centering
    \resizebox{1.0\textwidth}{!}{
    \begin{tabular}{l p{11.5cm}}
    \toprule
      Demographic Group & Terms \\
        \midrule
        Christian & christianize, christianese, Christians, christian-only, christianising, christiansand, christiany, jewish-christian, -christian, Christian., christianise, christianists, Christian, Christianity, christian-, Christians., christianity-, Christianity., christian-muslim, muslim-christian, christianized, christian-right, christianist, christian-jewish\\
         \midrule
        Jewish & judaïsme, jewish-canadian, half-jewish, part-jewish, anglo-jewish, jewes, french-jewish, -jewish, jewish-related, jewsish, christian-jewish, jewish-, jewish-zionist, anti-jewish, jewish-muslim, jewishgen, jews-, jewish-american, jewish., jewish-roman, jewish-german, jewish-christian, jewishness, american-jewish, jewsih, jewish-americans, jewish-catholic, jewish, jew-ish, spanish-jewish, semitic, black-jewish, jewish-palestinian, jewish-christians, jew, jewish-arab, jews, russian-jewish, jewish-owned, jew., german-jewish, judaism, jewishly, muslim-jewish, judaism., jewish-italian, jewish-born, all-jewish, austrian-jewish, catholic-jewish, jews., judaism-related, roman-jewish, jewish-themed, college-jewish, arab-jewish, jewish-only, british-jewish, judaisms, jewish-russian, pro-jewish, israeli-jewish, jewish-israeli \\
        \midrule
    \end{tabular}
    }
    \caption{The lists of words representing different demographic groups related to religion bias.}
    \label{tab:religion_word_list}
\end{table*}

\subsection{Age Bias}
\label{sec:age_words}

The terms associated with two different demographic groups for religion bias are listed in~\autoref{tab:age_word_list}.

\begin{table*}
    \centering
    \resizebox{1.0\textwidth}{!}{
    \begin{tabular}{l p{11.5cm}}
    \toprule
      Demographic Group & Terms \\
        \midrule
        young & adolescent, teen, teenage, teenaged, young, younger, twenty-year-old, 20-year-old, twentyfive-year-old, 25-year-old, thirty-year-old, 30-year-old, thirty-five-year-old, 35-year-old, forty-year-old, 40-year-old, twenty-something, thirty-something \\
        \midrule
        old & sixty-five-year-old, 65-year-old, seventy-year-old, 70-year-old, seventy-five-year-old, 75-year-old, eighty-year-old, 80-year-old, eighty-five-year-old, 85-year-old, ninety-year-old, 90-year-old, ninety-five-year-old, 95-year-old, seventy-something, eighty-something, ninety-something, octogenarian, nonagenarian, centenarian, older, old, elderly, retired, senior, seniorcitizen, young-at-heart, spry \\
        \midrule
    \end{tabular}
    }
    \caption{The lists of words representing different demographic groups related to religion bias.}
    \label{tab:age_word_list}
\end{table*}

\section{Social bias of the control experiment}
\label{sec:ap-control}
\autoref{tab:ap-temporal} and \autoref{tab:ap-random} show the social bias scores across time on the temporal corpora collected from X and the 5 subsets of corpus randomly sampled from a fixed time period, respectively.

\begin{table}[h]
    \centering
    {
    \begin{tabular}{ccc}
        \toprule
        Bias Scores & Female bias & Male bias \\
        \midrule
       Mar 2020 & 62.05 & 30.17  \\
Jun 2020 & 64.01 & 31.01 \\
Sep 2020 & 63.53 & 31.44 \\
Dec 2020 & 61.90 & 31.28 \\
Mar 2021 & 60.79 & 30.97 \\
Jun 2021 & 57.96 & 29.83  \\
Sep 2021 & 61.45 & 30.24 \\
Dec 2021 & 58.64 & 30.55 \\
Mar 2022 & 59.76 & 31.74 \\
Jun 2022 & 62.51 & 32.65 \\
Sep 2022 & 63.77 & 31.84 \\
         \midrule
    \end{tabular}
    }
    \caption{The social bias score of temporal corpora collected from X.}
    \label{tab:ap-temporal}
\end{table}

\begin{table}[h]
    \centering
    {
    \begin{tabular}{ccc}
        \toprule
        Bias Scores & Female bias & Male bias \\
        \midrule
         sample 1 &62.15  &59.89 \\
         sample 2 &62.36  &60.34  \\
         sample 3 &61.99  &60.19  \\
         sample 4 &62.36  &60.21  \\
         sample 5 &62.18  &59.96  \\
         \midrule
    \end{tabular}
    }
    \caption{The social bias score of 5 subsets of corpus randomly sampled from Jan to Mar 2020.}
    \label{tab:ap-random}
\end{table}

\autoref{tab:control-analysis} shows the standard deviation of social biases with different timestamps and within the same periods. 

\begin{table}[h]
    \centering
    \resizebox{0.48\textwidth}{!}
    {
    \begin{tabular}{ccc}
        \toprule
        Standard deviation & Female bias & Male bias \\
        \midrule
       across time & 2.03  & 0.84 \\
       same timestamp & 0.16 & 0.19 \\
         \midrule
    \end{tabular}
    }
    \caption{The standard deviations of temporal corpora collected from X and the subset of corpus random sampled from January to March 2020.}
    \label{tab:control-analysis}
\end{table}

\section{Results of COHABERT}

\subsection{Biases in COHABERT}
\label{sec:cohabert-bias}

\begin{table}[htb!]
    \centering
    \resizebox{0.48\textwidth}{!}
    {
    \begin{tabular}{lcccc}
    \toprule
       &mean & lower/upper & SE &SD  \\
        \midrule
        \textbf{CrowS-Pairs} & & & & \\
       OVERALL BIAS &47.83 & 46.59/49.23 & 0.79 &3.59 \\
       sexual-orientation &54.64 & 49.35/58.63 & 2.77 &12.74 \\
       disability &40.50 & 35.92/44.68 & 2.68 &12.32 \\
       socioeconomic &47.24 & 44.39/50.52 & 1.85 & 8.54 \\
       religion &38.38 & 35.81/42.05 & 1.83 &8.47 \\
       race-color &50.56 & 47.40/53.27 & 1.77 &8.21 \\
       appearance &47.46 & 44.52/50.16 & 1.73 &7.89 \\
       nationality &48.40 & 46.04/51.16 & 1.54 &7.02 \\
       age &48.16 & 46.04/50.69 & 1.40 &6.45 \\
       gender &45.74 & 44.86/46.75 & 0.58 &2.65 \\
       \midrule
        \textbf{StereoSet} & & & & \\
        OVERALL BIAS &49.94 & 49.54/50.34 & 0.24 &1.11 \\
       religion &57.15 & 54.18/59.75 & 1.68 &7.65 \\
       gender &47.86 & 46.49/49.27 & 0.85 &3.88 \\
       profession &50.69 & 49.89/51.51 & 0.49 &2.24 \\
       race &49.30 & 48.51/50.08 & 0.48 &2.20 \\
        \midrule
    \end{tabular}
    }
    \caption{The confidence interval and standard error computed using bootstrapping for each of the bias types on the CrowS-Pairs and StereoSet benchmarks for COHABERT models. SE and SD represent standard error and standard deviation, respectively. Lower/upper indicates the lower/upper bound of the confidence intervals. In each dataset, different bias types are sorted in the descending order of their SD.}
    \label{tab:cohabert-statistics-bias}
\end{table}

\begin{figure*}[h]
    \centering
    \subfigure[CrowS-Pairs]{
       
        \begin{minipage}[b]{0.95\linewidth}
        \includegraphics[width=0.9\linewidth]{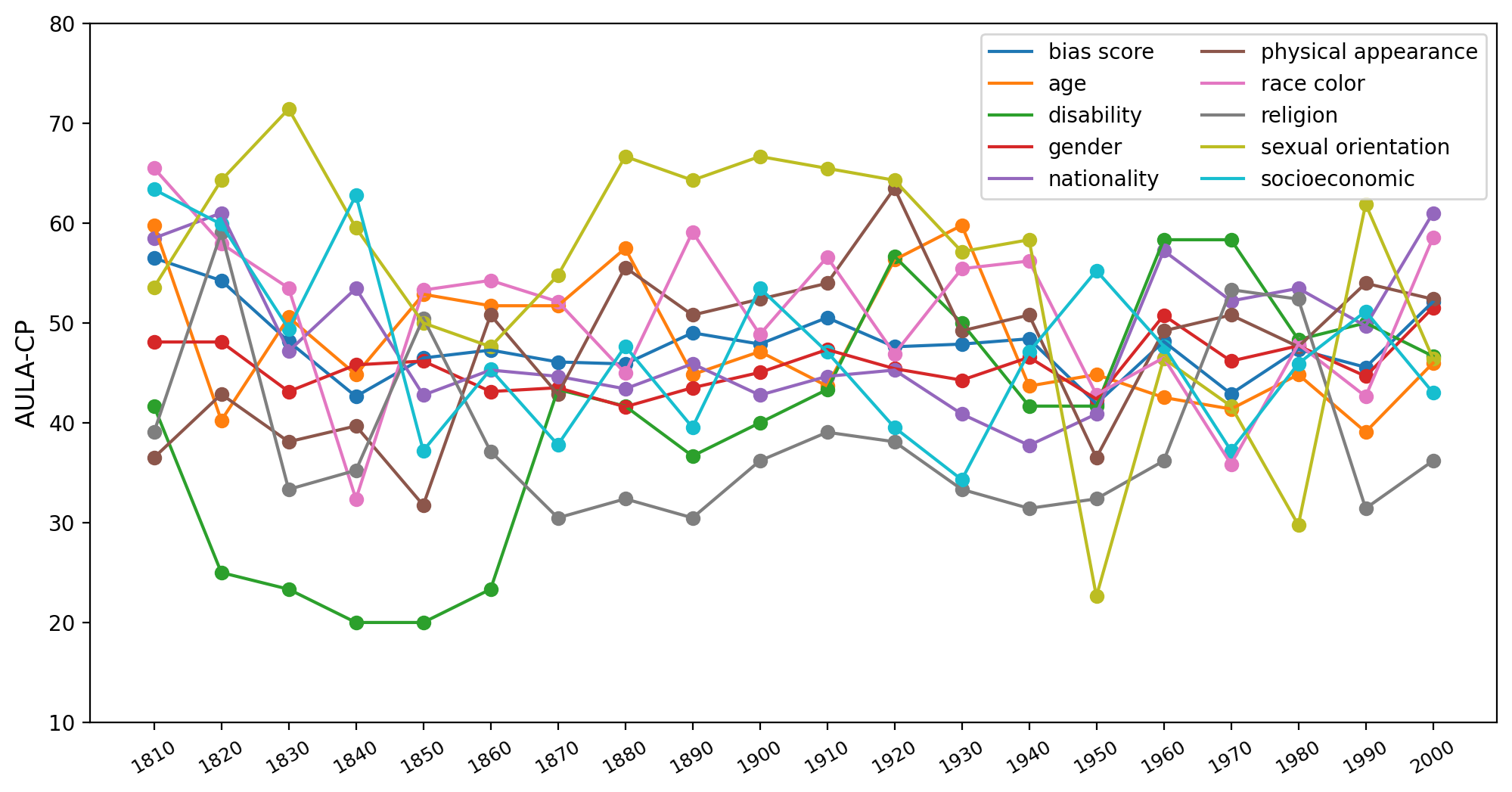}
        \label{fig:coha-aula-scores-cp}
        \end{minipage}
    }
    \subfigure[StereoSet]{
        \begin{minipage}[b]{0.95\linewidth} 
        \includegraphics[width=0.9\linewidth]{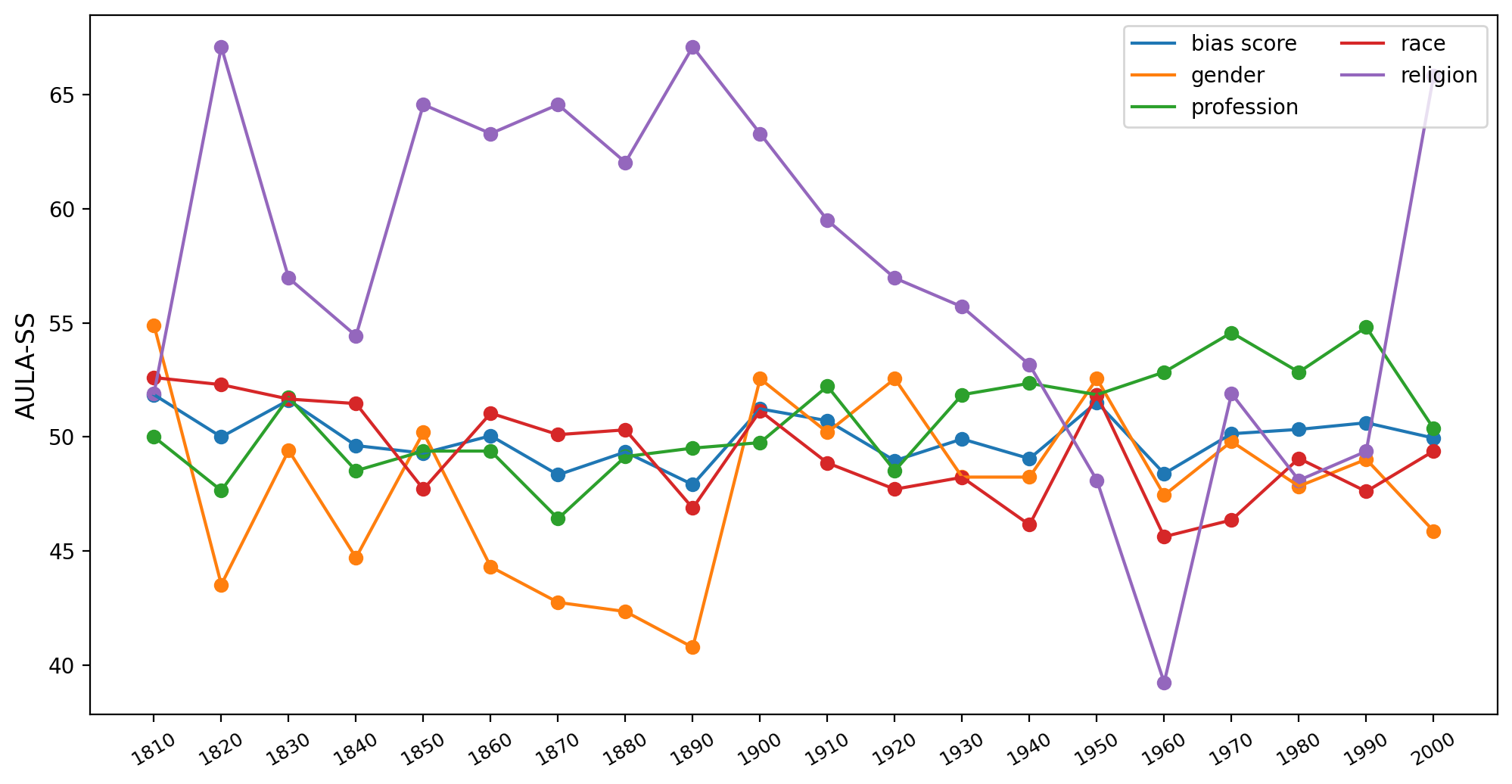}
       \label{fig:coha-aula-scores-ss}
        \end{minipage}
    }
    \caption{Social bias scores across time for different types of biases computed using the AULA metric for COHABERT models. Results evaluated on the CrowS-Pairs and StereoSet datasets are shown respectively on the top and bottom. The `bias score' (in dark blue) indicates the overall bias score.}
    \label{fig:cohabert-bias-score}
\end{figure*}

The result of the bias scores computed on both CrowS-Pairs and StereoSet for different bias types in COHABERT is shown in \autoref{fig:cohabert-bias-score}. 
The average and standard deviations are computed based on the AULA bias scores covering a period of 190 years, specifically from 1810 to 2000, with scores provided for each decade.

\subsection{Statistical Indicators of Bias Fluctuation Changes in COHABERT}
\label{sec:cohabert-statistics-bias}
The statistical indicators of bias fluctuation changes in COHABERT models are shown in \autoref{tab:cohabert-statistics-bias}.

\end{document}